\documentclass{article} 
\usepackage{amssymb}
\usepackage{multirow}
\usepackage{enumitem}
\usepackage{wrapfig}
\usepackage[final]{colm2025_conference_modified}

\usepackage{microtype}
\usepackage{hyperref}
\usepackage{url}
\usepackage{booktabs}

\usepackage{lineno}

\usepackage{xcolor}

\newcommand{\prover}{Goedel-Prover}

\newcommand{\dsprover}{DeepSeek-Prover-V1.5}

\newcommand{\internlmstep}{InternLM2.5-Step-Prover}

\newcommand{\internlmmath}{InternLM-Math}

\newcommand{\qwen}{Qwen2.5-Coder-32B}

\newcommand{\qweninstruct}{Qwen2.5-72B-Instruct}

\newcommand{\sonnet}{Claude-sonnet-3.5}

\newcommand{\lwb}{Lean Workbook}

\newcommand{\putnam}{PutnamBench}

\newcommand{\mathlib}{Mathlib4}

\newcommand{\miniff}{miniF2F}

\newcommand{\proofnet}{ProofNet}

\usepackage{microtype}
\usepackage{hyperref}
\usepackage{url}
\usepackage{booktabs}
\usepackage[most]{tcolorbox}
\usepackage{listings}

\definecolor{darkblue}{rgb}{0, 0, 0.5}
\hypersetup{colorlinks=true, citecolor=darkblue, linkcolor=darkblue, urlcolor=darkblue}

\title{{\prover}: A Frontier Model for Open-Source Automated Theorem Proving}

\author{
Yong Lin\thanks{YL and ST contribute equally to this work. \texttt{\{yong.lin,shangetang\}@princeton.edu}}~\thanks{Princeton Language and Intelligence, Princeton University.} 
 \\
\And
Shange Tang\footnotemark[1]~\footnotemark[2]
\And
Bohan Lyu\thanks{Tsinghua University.}\\
\And
Jiayun Wu\footnotemark[3]\\
\And
Hongzhou Lin\thanks{Amazon. This work is independent of and outside of the work at Amazon.} \\
\And
Kaiyu Yang\thanks{Meta FAIR. KY served an advisory role.
All experiments were conducted on PLI servers.} \\
\And
\quad Jia Li\thanks{Numina.} 
\quad
Mengzhou Xia\footnotemark[2]
\quad
Danqi Chen\footnotemark[2]
\quad
Sanjeev Arora\footnotemark[2]
\quad
Chi Jin\footnotemark[2]\\
}

\begin{document}

\ifcolmsubmission
\linenumbers
\fi

\maketitle

\begin{abstract}
We introduce Goedel-Prover, an open-source language model that achieves state-of-the-art (as of April 5 2025) performance in automated formal proof generation for mathematical problems. 
A key challenge in this field is the scarcity of formalized mathematical statements and proofs, which we address through the following approaches.
First, we train LLMs to convert natural language math problems from the Numina dataset  to equivalent formal statements in Lean 4.
This process creates the dataset Goedel-Pset-v1, which includes 1.64 million formal statements. Next, we develop a large dataset of formal proofs by training a series of provers. Each new prover can prove many statements that previous ones could not, and these new proofs are added to the training set for the next prover. Finally, we obtain the dataset Goedel-Pset-v1-solved, which contains proofs for over 800K statements from Goedel-Pset-v1.
Supervised fine-tuning (SFT) of \dsprover{-Base} on Goedel-Pset-v1-solved (i.e., no RL) yields a model that achieves a success rate of 57.6\% (Pass@32) on  \miniff{} benchmark, surpassing the previous leader \dsprover{} (trained using SFT + RL on a proprietary dataset) by 7.6\%.
On \putnam, \prover{}-SFT successfully solves 7 problems (Pass@512), ranking first on the leaderboard.  
Further RL training (including DPO) improves \prover-SFT's success rate to over 60\% (Pass@32) on \miniff{}.

To aid future research, we provide extensive discussion of our training methodology and design choices. We also fully open-source our codes, models, and datasets. Additionally, we open-source  formal proofs for $29.7$K  problems in \lwb{}, nearly doubling the $15.7$K  solved by prior provers. 
\end{abstract}

\section{Introduction}

\begin{figure}[h]
    \centering
\includegraphics[width=0.40\linewidth]{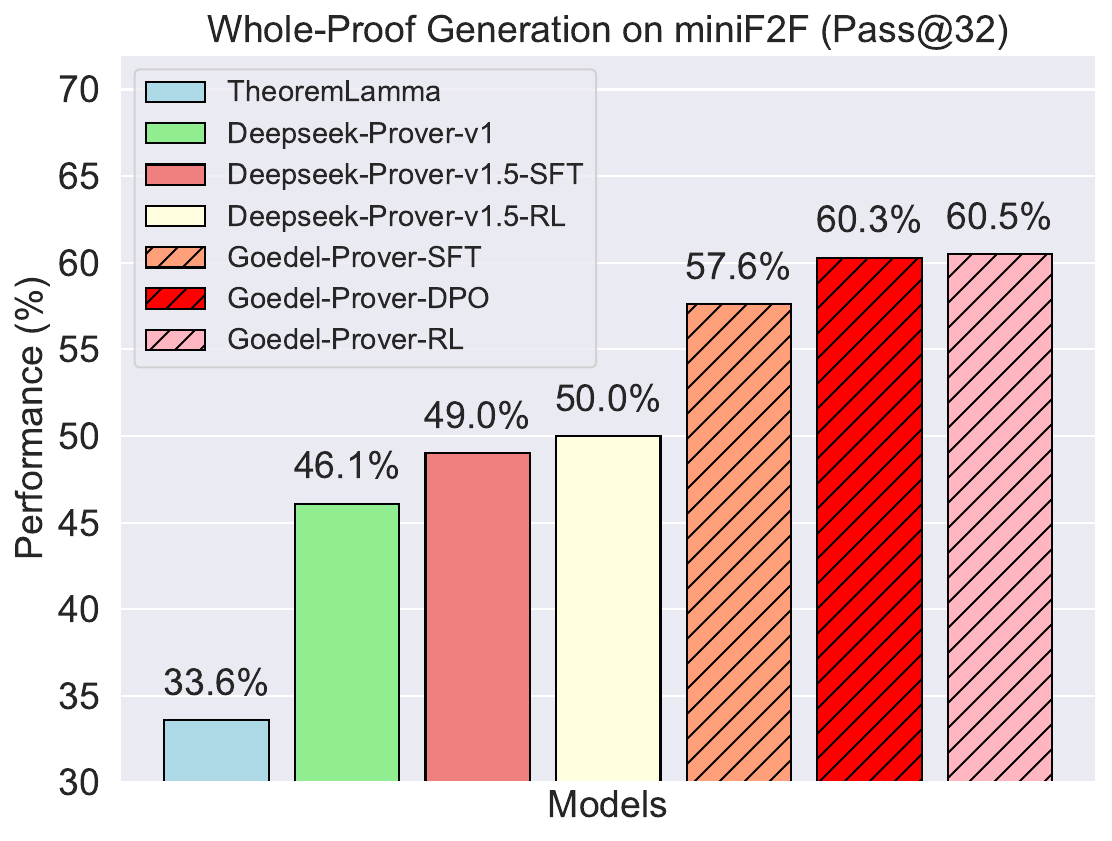}
\includegraphics[width=0.30\linewidth]{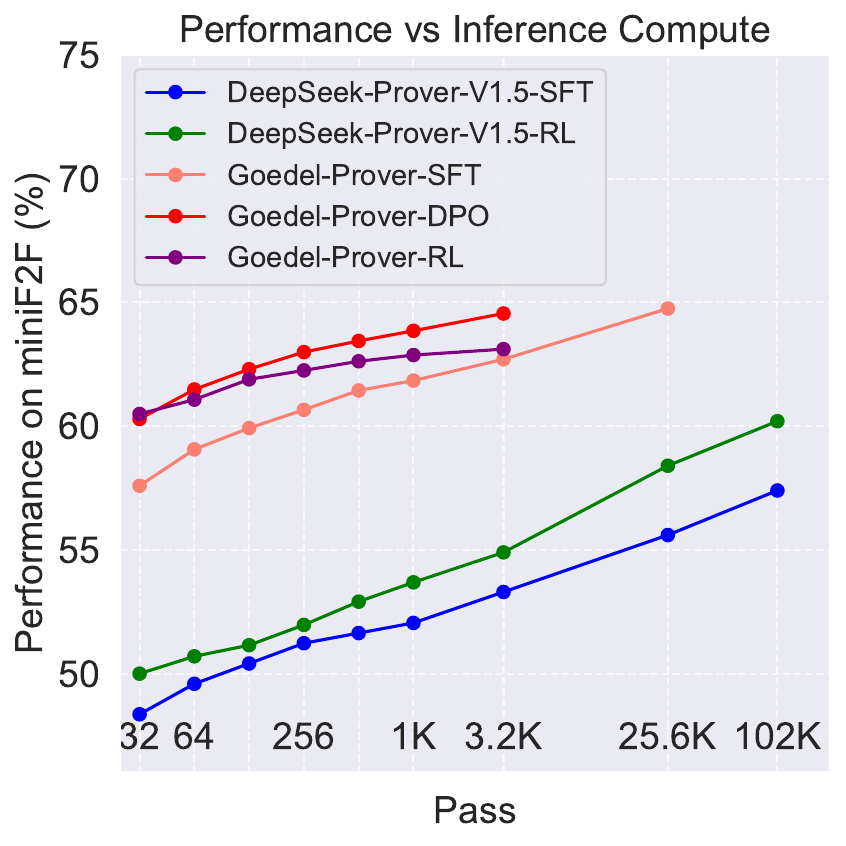}
\includegraphics[width=0.235\linewidth]{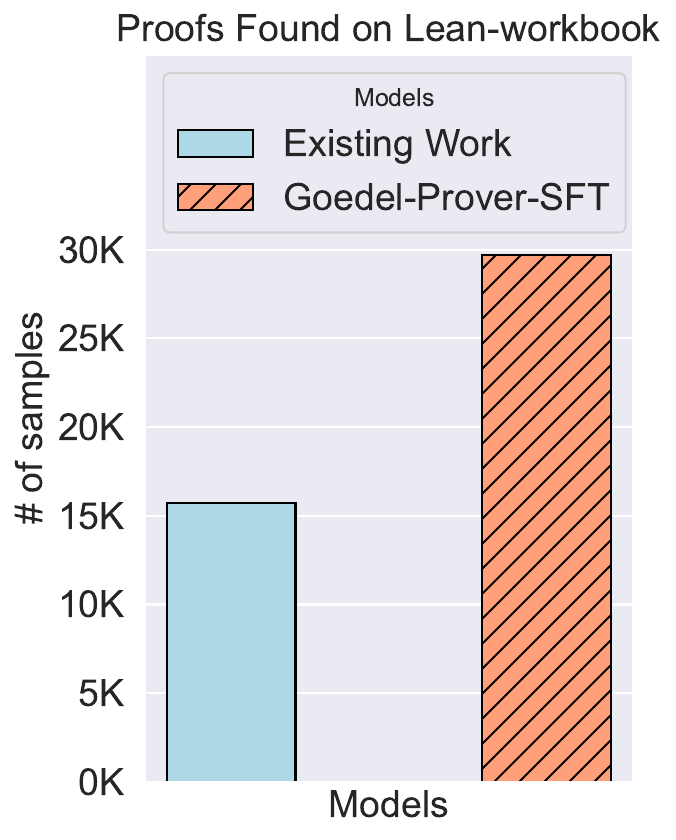}
    \caption{
    (\textbf{Left}) Pass@32 performance on miniF2F for whole-proof generation, compared to previous SOTA models.
    (\textbf{Middle}) A comparison of \prover and \dsprover{} on miniF2F performance across varying inference budgets, ranging from Pass@32, 64, 128, ..., to $4 \times 6400$.
    (\textbf{Right}) \prover-SFT solves 29.7K problems in the Lean Workbook. In comparison, \internlmstep~\citep{wu2024internlm2} and \internlmmath-Plus~\citep{ying2024internlm} collectively solved 15.7K samples.
    } 
    \label{fig:main_Results}
\end{figure}

Recent advancements in large language models (LLMs) have demonstrated remarkable capabilities in reasoning tasks, especially in solving mathematical problems \citep{, guo2025deepseek, yang2024qwen2}.
These models excel at reasoning through natural language, which we refer to \emph{informal reasoning}. However, natural language-based reasoning is difficult to automatically verify by machines, which undermines the reliability of informal reasoning in practical applications. This also makes it more difficult to further improve the reasoning capabilities of language models. In contrast to informal reasoning, \emph{formal reasoning} allows reasoning in a machine-verifiable format, opening up new possibilities for verification and automation. In particular, proof assistants such as Lean \citep{de2015lean, moura2021lean},  Isabelle \citep{paulson1994isabelle}, and Coq \citep{barras1997coq} provide formal languages that can express reasoning in a way that can be mechanically verified. Thus, it is of great interest to train LLMs to write proofs in these formal languages.

A significant challenge in training LLMs for theorem proving in formal languages is the scarcity of formalized math statements and proofs. Writing proofs for theorems expressed in formal languages is highly demanding and necessitates considerable domain expertise. Therefore, existing publicly available datasets for formal languages are limited in size. For example, the \lwb{} (including \lwb{} Plus) dataset \citep{ying2024lean, wu2024internlm2} comprises a total of 140K formal statements, where formal statements refer to problem statements in Lean without proofs.
However, only 15.7K of these statements come with formal proofs, which were found by InternLM2.5-StepProver and InternLM-Math-Plus \citep{ying2024lean, wu2024internlm2, ying2024internlm}. 
Additionally, the Open Bootstrapped Theorems dataset \citep{wang2024theoremllama} includes 107K statements with proofs sourced from \mathlib{} \citep{mathlib4}. However, \mathlib{} exhibits significant distribution shift from general problem-solving benchmarks, such as the widely used miniF2F~\citep{zheng2021minif2f}. See Section \ref{sec:design_choices} for detail.

In contrast to the scarcity of data in formal languages, there is a vast amount of math problems and solutions written in informal language. For example, Numina~\citep{li2024numinamath} includes 860K high-quality question and answer pairs sourced from MATH~\citep{hendrycks2021measuring}, GSM8K~\citep{cobbe2021training}, AMC~\citep{aops_wiki}, AIME~\citep{maa_aime_2024}, the AoPS Forum~\citep{aops_wiki}, Chinese K-12 Exams~\citep{shao2024deepseekmath}, World Olympiads, and synthetic data~\citep{mitra2024orca}. We start by training LLMs to formalize the problem statements in this dataset into Lean. To increase the diversity of the formalization styles, we train two formalizers. One is trained on informal and formal (I-F) statement pairs from \lwb{}, while the other is trained on I-F statement pairs annotated by Claude-sonnet-3.5~\citep{anthropic2024a}. We use these two formalizers to formalize the statements and then employ LLMs to ensure that the formal statements preserve the content of the informal statements.  Our efforts result in 1.64 million formal statements.

Using this extensive dataset of formal statements, we employ expert iteration \citep{polu2022formal} to train the prover to generate proofs. Notably, we train a model to generate complete proofs solely based on statements, without interacting with the Lean compiler during the generation process. This approach is referred to as the whole-proof
generation method~\citep{jiang2022draft, wang2024theoremllama, xin2024deepseek, xin2024deepseekv15}.
At the beginning of the expert iteration, we generate 16 proof candidates using \dsprover-RL (the previous SOTA) for each formal statement in our dataset, and then we verify the correctness of each candidate using Lean compiler. The correct proofs are then collected to train our iter-1 prover based on \dsprover-Base.
In subsequent rounds, we utilize our iter-\( k \) prover to collect new proofs and add them to the training data. We then perform supervised fine-tuning starting from \dsprover-Base for another round, resulting in the iter-\( (k+1) \) prover. We conduct a total of 8 iterations and observe a consistent improvement starting from the first iteration.

We demonstrate that expert iteration, with large-scale formalized statements can lead to SOTA performance in formal proof generation. Specifically,
\begin{itemize}[leftmargin=15pt, topsep=0pt]
\item Our model, \prover-SFT, outperforms \dsprover-RL (the previous SOTA model) by 7.6\% on miniF2F, achieving a Pass@32 score of 57.6\% (i.e., the prover generated 32 proofs for a problem, and 57.6\% of the problems have at least one correct proof verified by Lean) compared to \dsprover-RL's 50.0\%, as shown in Figure~\ref{fig:main_Results} (left). It consistently surpasses \dsprover-RL across all sampling budgets, including Pass@32, 64, and up to 25600, as shown in Figure~\ref{fig:main_Results} (middle).
\item We have cumulatively solved 29.7K problems in \lwb{}, significantly increasing the existing 15.7K proofs found by InternLM2.5-StepProver and InternLM-Math-Plus \citep{wu2024internlm2, ying2024internlm}, as shown in Figure~\ref{fig:main_Results} (right).
\item \prover-SFT solves 7 problems on PutnamBench by Pass@512\footnote{We initially solved 8 problems on PutnamBench. However, after discussing with the authors of PutnamBench, we discovered that one of the problems was mis-formalized. Therefore, this problem is not included in our count, and we report a total of 7 problems here.}, securing the {\textbf{\#1}} position on the leaderboard (Table~\ref{tab:putnambench}).  
\item \ifcolmsubmission
We open source our code, model, and the new proofs discovered in the \lwb{}, which will be released upon publication to facilitate future research. Additionally, we are going to open source our 1.64M formalized statements.
\else
We open source our codes\footnote{\url{https://github.com/Goedel-LM/Goedel-Prover}}, models\footnote{\url{https://huggingface.co/Goedel-LM/Goedel-Prover-SFT}} \footnote{\url{https://huggingface.co/Goedel-LM/Goedel-Prover-DPO}} \footnote{\url{https://huggingface.co/Goedel-LM/Goedel-Formalizer-32B-SonnetAnnotated}} \footnote{\url{https://huggingface.co/Goedel-LM/Goedel-Formalizer-32B-LeanWorkbookAnnotated}}, datasets\footnote{\url{https://huggingface.co/datasets/Goedel-LM/Goedel-Pset-v1}} \footnote{\url{https://huggingface.co/datasets/Goedel-LM/Goedel-Pset-v1-solved}}, and the new proofs discovered\footnote{\url{https://huggingface.co/datasets/Goedel-LM/Lean-workbook-proofs}} in the \lwb{} to facilitate future research.
\fi
\end{itemize}

To understand the factors behind \prover{}’s strong performance, we provide an in-depth discussion of our training recipe, analyzing the effects of scaling up training data, the diversity introduced by autoformalization, correlations among datasets, and alternative data synthesis strategies. Furthermore, although our final model is trained purely through supervised fine-tuning, we also explore direct preference optimization (DPO) and reinforcement learning (RL) techniques built on top of it. Our \prover-DPO and \prover-RL achieves a success rate over 60\% (Pass@32) on \miniff{}. However, we also find that DPO and RL-trained models tend to overfit to “shortcuts” and benefit less from increased inference-time compute.

\section{Related Work}
\paragraph{Automated theorem proving.} 
Automated theorem proving (ATP) is a long-standing problem in symbolic AI~\citep{robinson2001handbook}. Traditional approaches represent theorems in first-order logic and prove them using decision procedures~\citep{de2008z3,barbosa2022cvc5} and search~\citep{kovacs2013first,schulz2019faster}. The proof search has been enhanced by replacing handcrafted heuristics with machine learning techniques~\citep{urban2011malecop,kaliszyk2018reinforcement}. However, approaches based on first-order logic struggle to scale to complex theorems and often do not yield human-readable proofs.

In recent years, learning-based theorem proving has undergone a significant transformation. A notable approach, introduced by \citet{polu2020generative}, involves leveraging large language models to assist in theorem proving with proof assistants such as Lean~\citep{de2015lean, moura2021lean} and Isabelle~\citep{paulson1994isabelle}. Follow-up research has explored various avenues, such as retrieving useful lemmas~\citep{irving2016deepmath,mikula2024magnushammer,yang2024leandojo}, utilizing Monte Carlo tree search for proof discovery~\citep{lample2022hypertree}, and harnessing the capabilities of large language models (LLMs) for natural language reasoning~\citep{jiang2022draft,lin2024lean}. Notably, \cite{polu2023formal} was the first to apply expert iteration~\citep{anthony2017thinking} to theorem proving. This method alternates between two phases: (1) attempting to prove unsolved theorems and (2) enhancing the prover by incorporating newly discovered proofs into its training data. Expert iteration has yielded significant improvements in several recent provers~\citep{wu2024internlm2,xin2024deepseekv15}, including our {\prover}. 

Most automated theorem provers operate in a stepwise manner, generating individual proof steps that are then assembled into complete proofs using proof search algorithms. Recently, researchers have shown that generating entire proofs is feasible~\citep{first2023baldur,xin2024deepseek,wang2024theoremllama}. This approach avoids the costly search process, resulting in lower latency and potentially offering a more efficient use of computational resources during testing. While {\prover} also generates whole proofs, our data and methodology can, in principle, be adapted to develop stepwise provers as well.

\paragraph{Autoformalization and synthetic data generation.} 
The shortage of high-quality formal mathematical data poses a significant bottleneck in training theorem-proving models. While techniques like reinforcement learning may reduce the reliance on human-written proofs~\citep{alphaproof}, there remains a need for a substantial number of formal theorem statements. A promising approach is to synthesize formal statements through autoformalization, where large language models (LLMs) translate informal mathematical statements into formal ones~\citep{wu2022autoformalization,wu2024internlm2,xin2024deepseek,xin2024deepseekv15}. 

DeepSeek-Prover~\citep{xin2024deepseek} and InternLM2.5-StepProver~\citep{wu2024internlm2} have successfully implemented this strategy to formalize a large volume of statements into Lean for expert iteration. We adopt a similar approach. The difference is: while \citet{liu2024deepseek} focuses on formalizing their internal dataset, we concentrate on formalizing the Numina dataset~\citep{li2024numinamath} alongside a privately collected dataset. Additionally, we train two formalizers to enhance the diversity of formalization styles, which we demonstrate to be beneficial in Section~\ref{sect:results}. 


\section{Method}

We begin by translating informal statements (expressed in natural language) into formal statements (represented in Lean). Using these formal statements, we iteratively train our prover with proofs generated by the prover and verified by the Lean compiler. The details of each step are elaborated in the following parts.

\subsection{Statement Formalization}
\label{sect:statement_formalization}

\begin{figure}[t]
    \centering
    \includegraphics[width=0.45\linewidth]{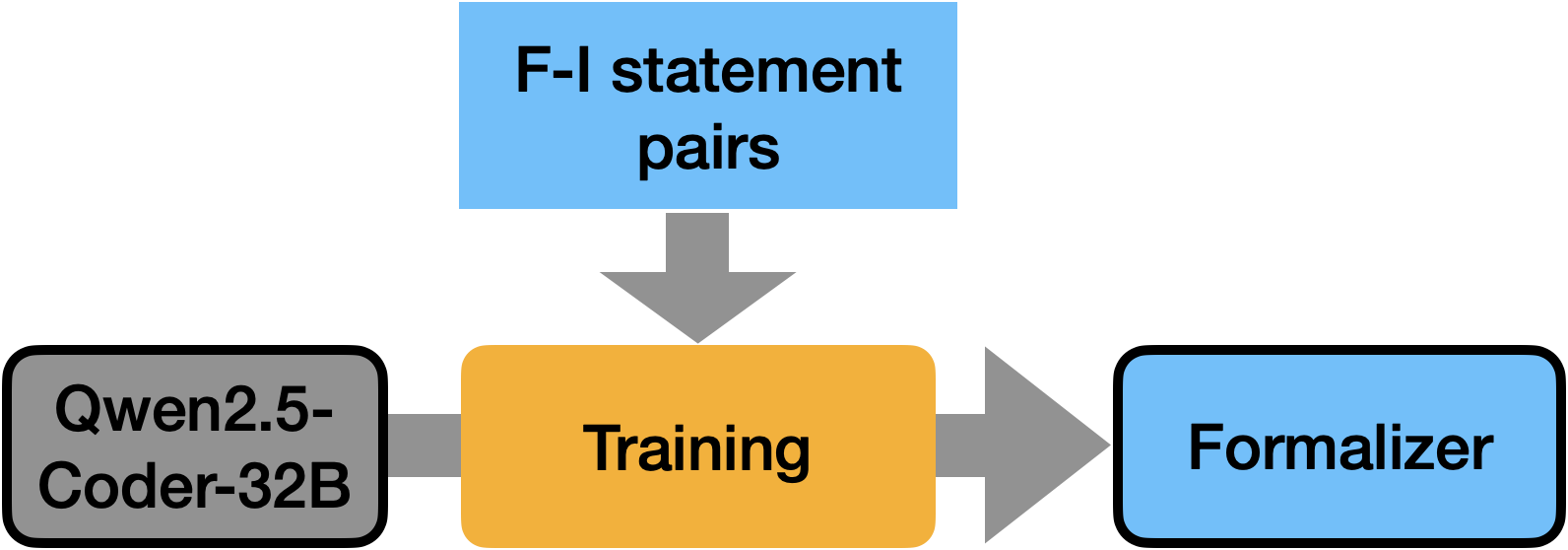}
    \includegraphics[width=0.48\linewidth]{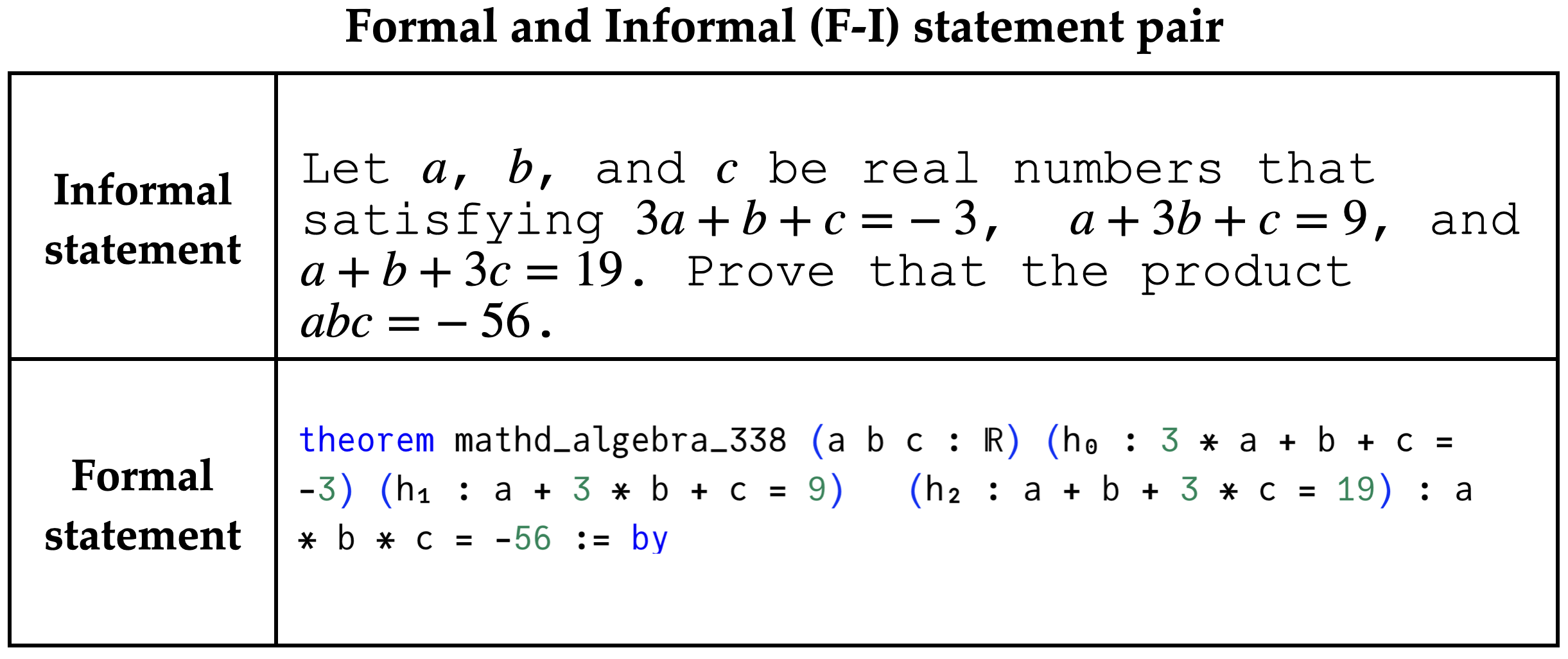}    
    \caption{This figure illustrates the training of the formalizers. The term ``F-I statement pairs" refers to pairs consisting of Formal and Informal (F-I) statements. An example is shown on the right part. We train two formalizers, Formalizer A and B, using F-I statement pairs sourced from various origins.}
    \label{fig:formalizer_training}
\end{figure}

We first train the statement formalizers to translate informal statements in Numina into formal statements as shown in Figure~\ref{fig:formalizer_training}.  
To enhance the diversity of formalized statements, we train two models to formalize informal statements. 

\begin{itemize}[leftmargin=15pt]
    \item \textbf{Formalizer A:} We train the Formalizer A model using Formal and Informal (F-I) statement pairs sourced from \lwb{}.
    \item \textbf{Formalizer B:} We employ \sonnet{} to formalize 230K statements from Numina. From this set, we extract 170K statements that successfully passed Lean compilation. These 170K F-I statement pairs are then used to train Formalizer B.
\end{itemize}
Both Formalizer A and B are trained using supervised fine-tuning with \qwen{}\footnote{\url{https://huggingface.co/Qwen/Qwen2.5-Coder-32B}}. 
The training of these two formalizers takes less than 24 hours on 8 H100 GPUs. See Appendix \ref{app:formalizer_examples} for examples of formalized statements form two formalizers, where we observe that even for the same problem, the style of the formalized statement can impact the prover's performance.


\begin{figure}[t]
    \centering
    \includegraphics[width=0.75\linewidth]{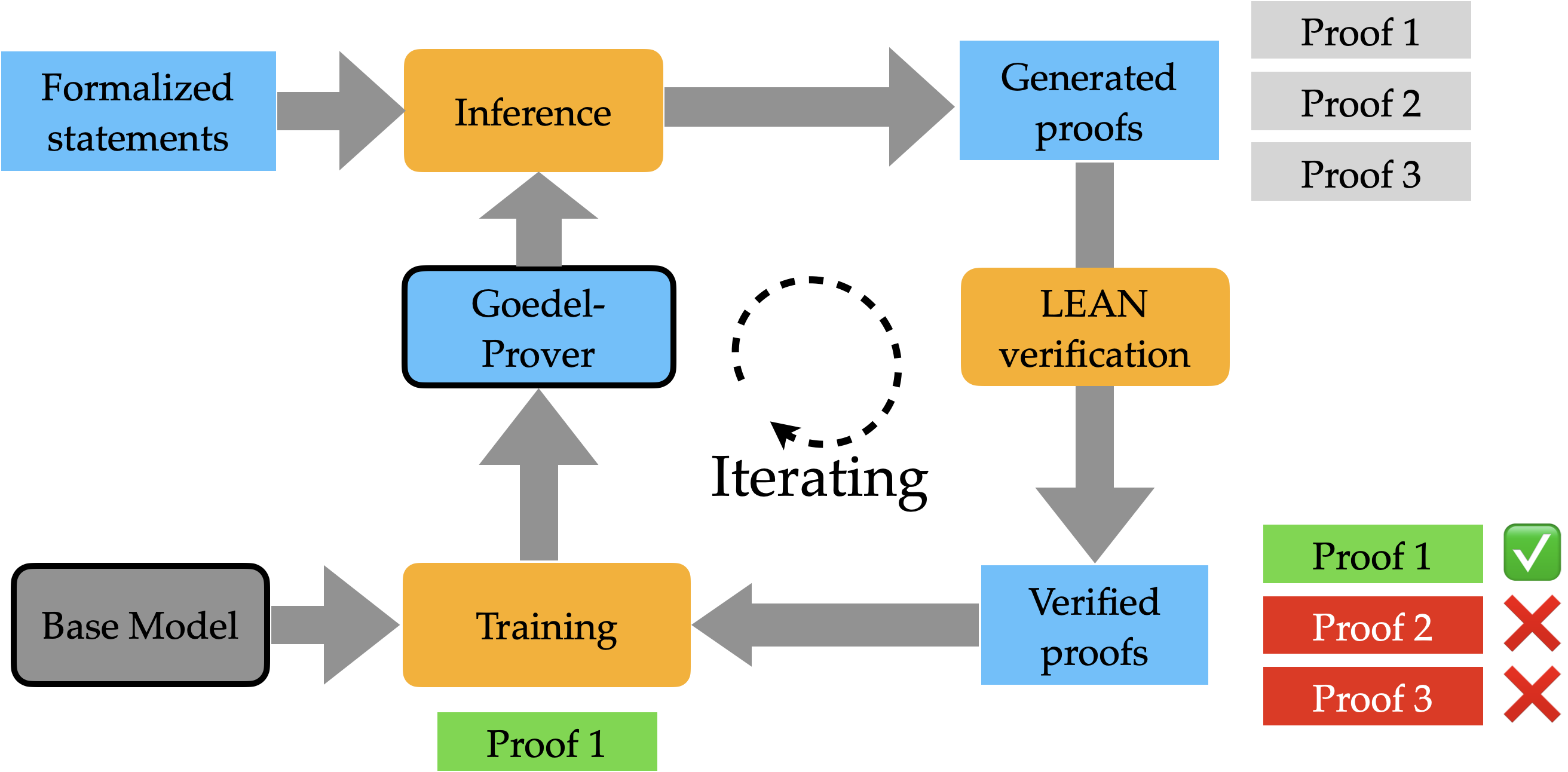}
    \caption{This figure illustrates the process of expert iteration. Each time, we utilize our iter-\( k \) prover to collect new proofs and add them to the training data. We then conduct supervised fine-tuning starting from \dsprover-Base for another round, resulting in the iter-\( (k+1) \) prover.}
    \label{fig:iterative_procedure}
\end{figure}

\paragraph{Quality assessment.}
We employ two tests to assess the quality of the formalized statements. First, the formalized statement must conform to Lean syntax and can successfully compile, with the potential proof replaced by the placeholder ``:= by sorry". This syntax check is known as the Compiling Correctness (CC) Test in the literature \citep{ying2024lean}. Second, the formalized statement must accurately capture the original informal problem, incorporating all assumptions, conditions, and implicit definitions. We refer to this second test as the Faithfulness and Completeness (FC) Test. For the FC test, we use \qweninstruct{}\footnote{\url{https://huggingface.co/Qwen/Qwen2.5-72B-Instruct}}, details are presented in Appendix \ref{app:qual_assess}.

In addition to formalizing the 860K open-sourced Numina~\citep{li2024numinamath} datasets, we also formalize a private 68K collection of math problems from Art of Problem Solving (AOPS), which has been collected and processed by the Numina group~\citep{li2024numinamath}. Out of a total of 928K informal statements, 760K have two valid formalized statements generated by Formalizer A and B, while 123K contain only one valid formalized statement.  After formalizing both the Numina and AOPS datasets, we further incorporate 140K statements from \lwb{}, including \lwb{ Plus}. As a result, we have a total of 1.78M formal statements.

\subsection{Expert Iteration}
\label{sect:iterative}
After obtaining a large collection of formalized statements in Section~\ref{sect:statement_formalization}, we employ expert iteration to train the prover~\citep{liu2024deepseek, wu2024internlm2, li2024hunyuanprover}, which is illustrated in Figure~\ref{fig:iterative_procedure}. Specifically, we first utilize \dsprover{-RL}\footnote{\url{https://huggingface.co/deepseek-ai/DeepSeek-Prover-V1.5-RL}} to generate 16 proofs for each statement. We then verify these proofs with the Lean compiler. If at least one proof solves the statement, we retain one proof per statement. In cases where multiple proofs are available, we randomly sample one solution. These collected proofs are used for supervised fine-tuning (SFT) based on \dsprover{-Base}\footnote{\url{https://huggingface.co/deepseek-ai/DeepSeek-Prover-V1.5-Base}}, resulting in the iter-1 prover. We continue this expert iteration process; each time, we use the iter-$k$ prover to generate answers and cumulatively collect correct solutions to train \dsprover{-Base} for the next iteration, the iter-$(k+1)$ prover. Refer to Appendix~\ref{app:training_details} for more details on each iteration.

We experiment with learning rates of \(1 \times 10^{-4}\) and \(5 \times 10^{-5}\), training for either 1 or 2 epochs.  We use the packing trick~\citep{tunstall2022natural} with a small batch size of 8 to speed up the training.  In each iteration, the training time for 1 epoch is approximately 12 hours using 4 H100 GPUs. The inference time for the 1.78M statements set by Pass@16 is 6 hours, utilizing 64 H100 GPUs. Additionally, the verification time for these proofs requires 10 hours with 8,000 CPUs.

\section{Results}
\label{sect:results}
\paragraph{Benchmarks.} Following the works of \citep{wang2024theoremllama, xin2024deepseek, wu2024internlm2, li2024hunyuanprover}, we primarily use miniF2F \citep{zheng2021minif2f} as our main evaluation benchmark. We also track the problems solved by our prover in \lwb{}~\citep{ying2024lean} and investigate the performance on \proofnet~\citep{azerbayev2023proofnet} and \putnam{}~\citep{tsoukalas2024putnambench}. Additionally, we uniformly sample a subset from our formalized dataset to create a held-out evaluation dataset. Below, we provide descriptions of each dataset.
\begin{itemize}[leftmargin=15pt]
    \item \miniff{} \citep{zheng2021minif2f} is a formal theorem proving benchmark, consisting of 488 problem statements (244 validation and 244 test problems) in Lean. The problems are drawn from high-school exercises, as well as high-school level competitions including the AIME, AMC, and the International Mathematical Olympiad (IMO). The  original benchmark was released in Lean 3, and for our analysis, we use the version of \miniff ~in Lean 4.9.0 provided by \cite{xin2024deepseek}.
    \item \proofnet{} \citep{azerbayev2023proofnet} is a formal theorem proving benchmark of undergraduate-level mathematics, consisting of 371 problem statements in Lean (185 validation and 186 test problems). The problems are primarily drawn from undergraduate pure mathematics textbooks, covering topics such as real and complex analysis, linear algebra, abstract algebra, and topology. The original benchmark was released in Lean 3, and for our analysis, we use the version of \proofnet{} in Lean 4.9.0 provided by \cite{xin2024deepseek}.
    \item \lwb{} \citep{ying2024lean} is a large-scale Lean 4 problem set formalized from natural language math problems (mainly from the forum AOPS), which consists of 140K statements in Lean 4. We also monitor the problems solved by our model during the expert iteration process. Notably, the problem set from \lwb{} is included in this training, which is consistent with \dsprover~\citep{xin2024deepseek} and InternLM2.5-StepProver~\citep{wu2024internlm2}.
    \item \putnam{} \citep{tsoukalas2024putnambench} is a formal theorem proving benchmark on competition mathematics problems sourced from the William Lowell Putnam Mathematical Competition years 1962 - 2023. \putnam comprises 644 Lean 4 statements, covering algebra, analysis, number theory, geometry, combinatorics, probability and set theory.
    \item NuminaTest. We randomly sample 250 statements from our formalized Numina dataset and use it as a held-out testing set. We refer to this subset as NuminaTest.
\end{itemize}

\begin{figure}[t]
    \centering
    \makebox[\textwidth][c]{   \includegraphics[width=0.26\linewidth]{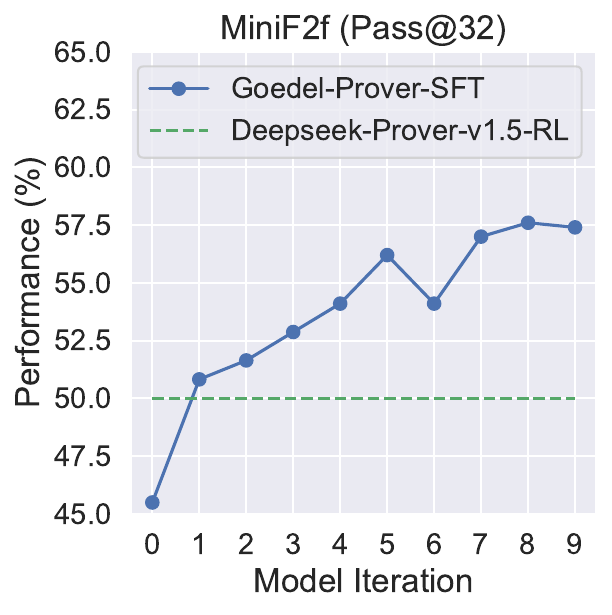}
    \includegraphics[width=0.26\linewidth]{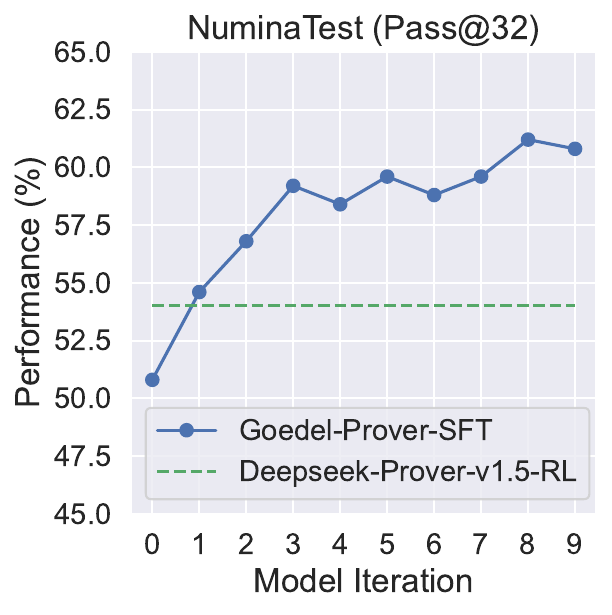}
    \includegraphics[width=0.26\linewidth]{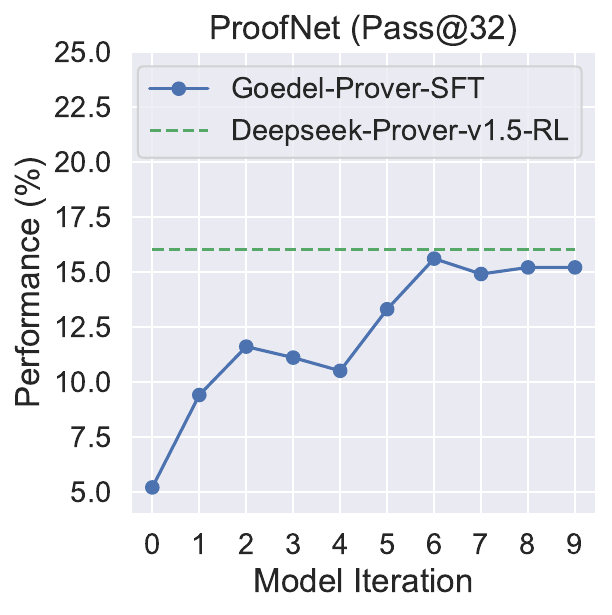}
    \includegraphics[width=0.26\linewidth]{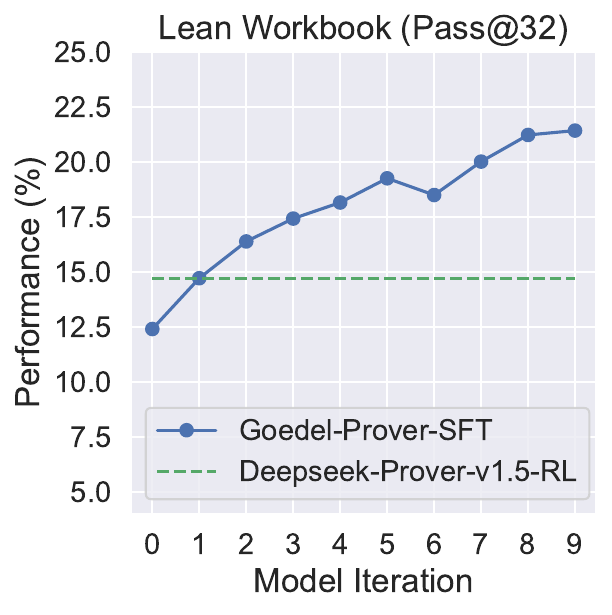}
    
    }
    \caption{The figures show the performance of our model on the four datasets at each iteration. We gradually increase the size of the problem set and add more training data. The details of each iteration are shown in Table~\ref{tab:data_for_training}. }\label{fig:iterative_training}
\end{figure}

\paragraph{Main results.} 
The performance on miniF2F is shown in Table~\ref{tab:full_proof_generation_comparison}. The Pass@32 performance of our \prover-SFT is 57.6\%, surpassing the previous SOTA open source model, \dsprover-RL, by 7.6\%. We observe that our \prover-SFT's Pass@32 is even better than \dsprover-RL's Pass@3200 by 2.7\%.  Furthermore, when both evaluated by Pass@3200, our model achieves 62.7\%, surpassing \dsprover-RL's 54.9\% by 7.8\%. Figure~\ref{fig:main_Results} illustrates the inference time scaling curve for our \prover-SFT, \dsprover-RL and \dsprover-SFT. \prover-SFT demonstrates significant improvements over both \dsprover-RL and \dsprover-SFT across all inference compute budgets. Figure~\ref{fig:iterative_training} illustrates the performance of our model during each iteration. Overall, we observe a relatively consistent improvement in performance across iterations. 

\begin{table}[t]
    \centering
    \resizebox{0.75\linewidth}{!}{
    \begin{tabular}{c|cc}
    \toprule
       Whole-Proof Generation Model & Pass & Performance \\
         \midrule
        TheoremLamma \citep{wang2024theoremllama} & 128 & 33.6\%\\
        Deepseek-Prover-V1 \citep{xin2024deepseek} & 32 & 46.1\% $\pm$ 0.5\% \\
        \dsprover-SFT \citep{xin2024deepseekv15} &32 &  48.2\% $\pm$ 0.6\%\\
        \dsprover-RL \citep{xin2024deepseekv15} &32 & 50.0\% $\pm$ 0.5\%\\
        {\prover-SFT}&32 & {\textbf{57.6\% $\pm$ 0.7\%}}\\        
        \midrule
        \dsprover-SFT \citep{xin2024deepseekv15} & 3200 & 53.3\%\\ 
        \dsprover-RL \citep{xin2024deepseekv15} &  3200 & 54.9\%\\ 
         {\prover-SFT}& 3200 & {\textbf{62.7\%}}\\
        \midrule
        \dsprover-SFT \citep{xin2024deepseekv15} &  $4 \times 6400$ & 55.8\%\\ 
        \dsprover-RL \citep{xin2024deepseekv15} &  $4 \times 6400$ & 58.5\%\\ 
         {\prover-SFT}&  $4 \times 6400$ & {\textbf{64.7\%}}\\
        \bottomrule
    \end{tabular}
    }
    \caption{Whole-proof generation performance on miniF2F.}
    \label{tab:full_proof_generation_comparison}
\end{table}

\paragraph{\putnam{} performance.} \prover-SFT solves 7 out of 644 problems in \putnam{} (Pass@512), achieving the first place on the \putnam{} leaderboard. The previous SOTA method ABEL~\citep{gloeckle24abel}  solves 7 with a slightly higher inference budget (Pass@596) and \internlmstep{}~\citep{wu2024internlm2} solves 6 (Pass@$2\times32\times600$).  The performance is summarized in Table~\ref{tab:putnambench}.

\begin{table}[t]
\centering
\resizebox{\linewidth}{!}{
\begin{tabular}{@{}clccc@{}}
\toprule
Ranking & Model & Type & Num-solved & Compute (Pass) \\ \midrule
1 & \textbf{\prover-SFT} \textcolor{green}{$\diamond$} & Whole-Proof Generation & 7 & 512 \\
1 & ABEL~\citep{gloeckle24abel}  & Tree Search Method & 7 & 596 \\
3 & \textbf{\prover-SFT} \textcolor{green}{$\diamond$} & Whole-Proof Generation & 6 & 32 \\
3 & InternLM2.5-StepProver~\citep{wu2024internlm2} \textcolor{green}{$\diamond$} & Tree Search  Method & 6 & 2$\times$32$\times$600 \\
5 & InternLM 7B~\citep{ying2024internlm} \textcolor{green}{$\diamond$} & Whole-Proof Generation & 4 & 4096 \\
6 & GPT-4o  & Whole-Proof Generation & 1 & 10 \\
7 & COPRA (GPT-4o)~\citep{thakur2023language}& Whole-Proof Generation & 1 & 1 \\
8 & ReProver w/ retrieval~\citep{yang2024leandojo} \textcolor{green}{$\diamond$} & Whole-Proof Generation & 0 & 1 \\
9 & ReProver w/o retrieval~\citep{yang2024leandojo} \textcolor{green}{$\diamond$} & Whole-Proof Generation & 0 & 1 \\ \bottomrule
\end{tabular}
}
\caption{Number of problems solved on \putnam{} statements (out of 644). \prover-SFT achieves the first place in the leaderboard. The performance numbers for existing works are taken from the leaderboard. Here \textcolor{green}{$\diamond$} inidicates open-source models.}
\label{tab:putnambench}
\end{table}




\paragraph{Proofs found in \lwb{}.} The \lwb{}, which includes \lwb{}-plus \citep{ying2024lean,wu2024internlm2}, formalizes 140K high-quality problems sourced from AOPS and the Compfiles data. Currently, proofs for only 15.7K statements in \lwb{} have been found and made open-source by InternLM2.5-StepProver \citep{wu2024internlm2} and InternLM-Math-Plus \citep{ying2024internlm}. In contrast, our model has discovered a significantly larger set of proofs within \lwb{}, cumulatively solving 29.7K problems, as shown in Figure~\ref{fig:main_Results} (right). We open-source all the proofs found by our model to benefit the research community.


\section{Dissecting the training recipe}
\label{sec:design_choices}

\paragraph{Scaling up the number of formal statements improves model performance.}
\begin{figure}
    \centering
    \includegraphics[width=0.4\linewidth]{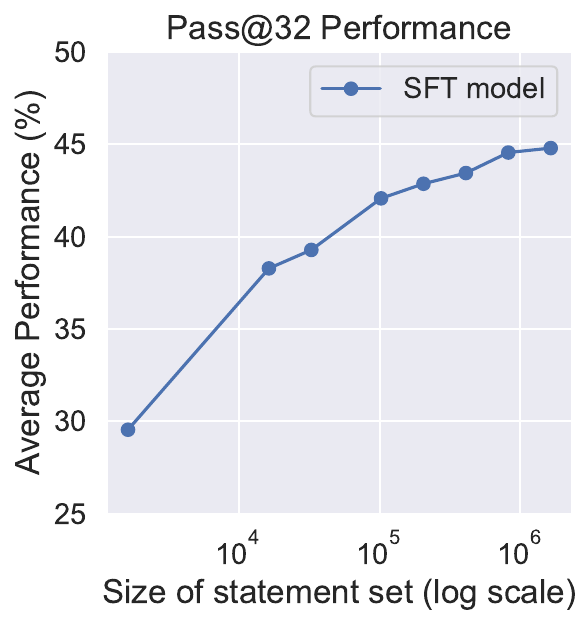}
    \caption{Model performance under different size of training statement set, illustrating the value of scale.}
    \label{fig:scale}
\end{figure}
Figure \ref{fig:scale} shows the performance of provers (average on \miniff{}, \proofnet{} and NuminaTest) trained on different sizes of the formal statement set.
For each statement, the corresponding proof is obtained using \prover-SFT. We observe a consistent improvement in model performance as the size of the statement set increases, underscoring the value of scale in training effective provers. 
\paragraph{Increasing the diversity of formalization styles is beneficial.}
Table~\ref{tab:formalization_style} presents the performance of iter-8 provers trained on different formalization styles of statements, with proofs generated by the iter-7 prover. We find that a prover trained on a mixture of styles—combining statements produced by both Formalizer A and Formalizer B—outperforms provers trained on a single formalization style. This result suggests that exposure to diverse formalization patterns improves the model’s generalization and reasoning ability.
\begin{table}[t]
    \centering
    \resizebox{0.8\linewidth}{!}{
    \begin{tabular}{c|ccc|c}
    \toprule
        Formalization Model  &  miniF2F &ProofNet & NuminaTest& Average\\
        \midrule
        Formalizer A only & 56.5\% & 13.8\%&59.6\% &43.3\%\\
        Formalizer B only & 56.2\% & \textbf{15.2}\% & 60.0\%&43.8\%\\
        \midrule
        Formalizer A and B & \textbf{57.6}\% &\textbf{15.2}\% & \textbf{61.2}\%&\textbf{44.7}\%\\
        \bottomrule
    \end{tabular}
    }
    \caption{An ablation study on using two formalizers to formalize the statements. Using statements formalized by both formalizers improves the model's performance, illustrating the value of diverse formalization styles.} 
    
    \label{tab:formalization_style}
\end{table}

\paragraph{Correlations among datasets.} 
\begin{figure}
    \centering
    \includegraphics[width=0.4\linewidth]{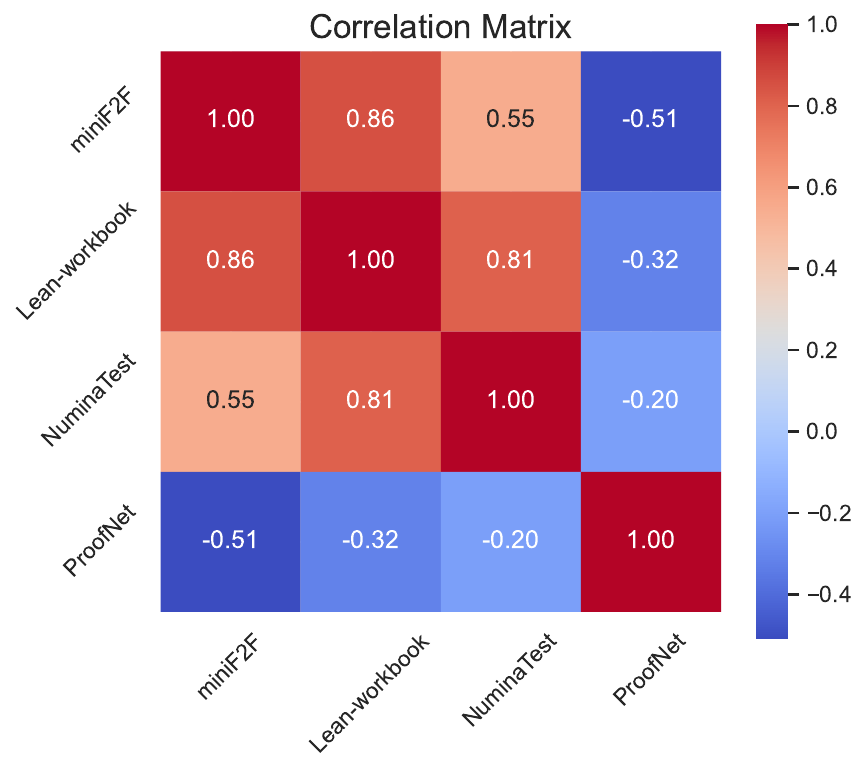}
    \caption{Correlation of model performance across datasets. \proofnet{} shows low correlation with the others.}
    \label{fig:proofnet_correlation}
\end{figure}
We evaluate model performance across different training iterations and hyperparameter settings, and compute the correlation of performance across multiple datasets (see Figure~\ref{fig:proofnet_correlation}). We observe that the model performance on \proofnet{} is negatively correlated with the performance on \miniff{}, \lwb{}, and NuminaTest.
Further more, we investigate the effect of including \mathlib{} in the training data. As shown in Table~\ref{tab:proofnet_perforemance}, incorporating \mathlib{} improves performance on \proofnet{} but leads to a performance drop on \miniff{}. These findings suggest a distribution shift between \proofnet{}/\mathlib{} and the other datasets. Specifically, \mathlib{} and \proofnet{} tend to focus on the manipulation of mathematical concepts, whereas datasets like \miniff{}, \lwb{}, and NuminaTest contain more Olympiad-style problems that emphasize complex reasoning over formal mathematical content. Illustrative examples are provided in Appendix~\ref{app:examples}. Despite the observed distribution shift, we continue to include \mathlib{} in the training set from the sixth iteration onward, following the approach of \dsprover-RL~\citep{xin2024deepseekv15} and TheoremLamma~\citep{wang2024theoremllama}, with the aim of enhancing the model’s general capability across a broader range of mathematical domains. Additional training details can be found in Appendix~\ref{app:training_details}.
\begin{table}[t]
    \centering
    \resizebox{0.95\linewidth}{!}{
    \begin{tabular}{@{}lccccc@{}}
        \toprule
        Model & Training Dataset &  miniF2F & ProofNet & NuminaTest  & Average \\ 
        \midrule
        Deepseek-RL & -- & 50.0\% & \textbf{16.0\%} & 53.6\% & 39.9\% \\ 
        \midrule
        Iter-6 prover & Iter-5 proofs &  \textbf{56.6\%} & 13.3\% & \textbf{59.2\%} & \textbf{43.0\%} \\ 
        Iter-6 prover  & Iter-5 proofs  + \mathlib{} & 54.1\% & 15.6\% & 58.8\% & 42.8\% \\ 
        \bottomrule
    \end{tabular}
    }
    \caption{Incorporating \mathlib{} into the training data enhances performance on \proofnet{} but reduces performance on \miniff{} and NuminaTest, suggesting distribution shift between \mathlib{}/\proofnet{} and other datasets.}
    \label{tab:proofnet_perforemance}
\end{table}

\paragraph{Alternative approach for data synthesis.} 
In addition to autoformalizing statements and use the prover to provide proofs, we also explored alternative strategies for constructing training datasets, focusing on solving difficult problems by a divide-and-conquer strategy. Inspired by \citet{jiang2022draft}, we implemented the following pipeline: (1) generate a proof for a formal statement using OpenAI’s o1-preview model, (2) extract a high-level ``sketch'' of the proof and (3) apply \dsprover-RL to prove the subgoals provided by the sketch. If all the subgoals are successfully completed, we obtain a valid proof for the original problem. Implementation details are provided in Appendix~\ref{app:sketch_proof}. However, this pipeline turned out to be ineffective in practice. When applied to the \miniff{} validation set (244 problems), it successfully solved only 76 problems—considerably fewer than the 158 problems solvable by \dsprover-RL alone. Moreover, out of the 76 problems solved, only one is not solved by \dsprover-RL, implying that the marginal gain from this pipeline is limited.

\paragraph{Exploring DPO and RL training.} We further explored DPO and RL training on top of \prover-SFT. We implemented offline Direct Preference Optimization (DPO) \citep{rafailov2023direct}
and online Group Relative Policy Optimization (GRPO) \citep{shao2024deepseekmath}, implementation details are provided in Appendix~\ref{app:RL}. Table \ref{tab:RL} shows that although DPO and GRPO improve the model's Pass@32 performance, the average proof length grows substantially, and the frequency of certain patterns increases sharply. This phenomenon indicates that the model is overfitting to some syntactic patterns or ``shortcuts'', which is related to ``reward hacking'' \citep{chen2024odin}. For example, the Lean tactic \texttt{try} allows trying a tactic and continue execution regardless of whether it works or not. Although often harmless—and occasionally useful—its overuse can lead to ineffective proofs and substantial verification costs. The RL-trained model begins to excessively favor this pattern, ultimately impairing its reasoning and generalization capabilities.

\begin{table}[t]
    \centering
    \resizebox{\linewidth}{!}{
    \begin{tabular}{l c c c c}
        \toprule
        Training method & \begin{tabular}{c} Pass@32 \\ (minif2f) \end{tabular} & \begin{tabular}{c} Pass@3200 \\ (minif2f) \end{tabular}  & \begin{tabular}{c} Average \\ proof length \end{tabular} & \begin{tabular}{c} Average number of \\ tactic ``\texttt{try}'' \end{tabular} \\
        \midrule
        \textbf{SFT} & 57.5\% & 62.7\% & 298 & 1.50 \\
        \textbf{DPO} & 60.3\% & 64.6\% & 486 & 10.89 \\
        \textbf{Length-penalized DPO}  & 59.8\% & 63.1\% & 308 & 1.11 \\
        \textbf{GRPO} & 60.5\% & 63.1\% & 355 & 5.16 \\
        \bottomrule
    \end{tabular}
    }
    \caption{Models' behavior under different training methods. RL methods show improvement on \miniff{} at Pass@32, but the improvement at Pass@3200 is limited . Furthermore, RL models are prone to excessively favor patterns such as \texttt{try}, which also causes the proof length to increase.}
    \label{tab:RL}
\end{table}
Further experiments show that adding a length penalty during DPO training helps reduce this overfitting. However, we observe that scaling up inference-time compute yields significantly smaller gains for models fine-tuned with either GRPO or length-penalized DPO, compared to the SFT model. As shown in Table~\ref{tab:RL}, these models achieve a 3\% improvement over \prover-SFT on Pass@32, but this gain diminishes when increasing inference-time compute—for example, at Pass@3200. This indicates that RL training may reduce output diversity, leading to less efficient inference-time scaling.

\section{Discussion}
Further discussions on the characteristics of proofs generated by \prover-SFT and potential areas for improvement are provided in Appendix \ref{app:discussion}.

\ifcolmsubmission
\else
\section*{Acknowledgments}
We thank Haoyu Zhao, Hubert Strauss and Suozhi Huang for their helpful discussions.
\fi


\bibliography{colm2025_conference}
\bibliographystyle{colm2025_conference}
\newpage
\appendix

\section*{Appendix}

\section{Statement Formalization Details}
\subsection{Examples of formalized statements}
\label{app:formalizer_examples}
Table \ref{tab:formalizer_comparison_two_columns} presents two examples in which both Formalizer A and Formalizer B yield reasonable formalizations. However, our final prover exhibits varying performance on these formalized statements, highlighting the influence of formalization style on model effectiveness.

\begin{table}[ht]
\centering
\renewcommand{\arraystretch}{1.5} 
\resizebox{\linewidth}{!}{
\begin{tabular}{|p{0.15\textwidth}|p{0.4\textwidth}|p{0.3\textwidth}|}
\hline
 & \textbf{Example 1} & \textbf{Example 2} \\ \hline

\textbf{Informal Statement} & 
The function $f(x) = 2^{|x|} + 3x^2 + ax + 1$ is an even function, then $a$ equals $a=0$. &
If $x$ and $\log_{10} x$ are real numbers and $\log_{10} x<0$, show that $0<x<1$.
 \\ \hline

\textbf{Formalizer A Output} &
\begin{minipage}[t]{\linewidth}
    \includegraphics[width=\linewidth]{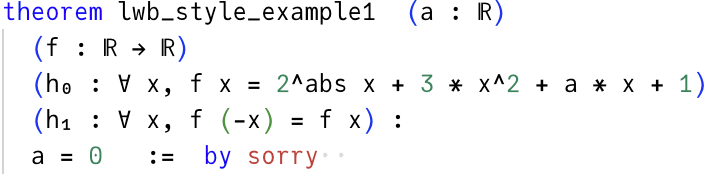}
\end{minipage} \newline
\texttt{Pass rate: 14/16} &
\begin{minipage}[t]{\linewidth}
    \includegraphics[width=0.8\linewidth]{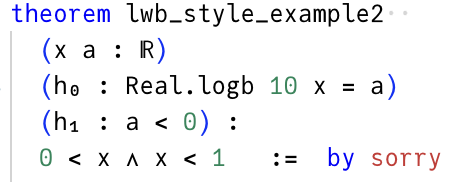}
\end{minipage} \newline
\texttt{Pass rate: 0/16} \\ \hline

\textbf{Formalizer B Output} &
\begin{minipage}[t]{\linewidth}
    \includegraphics[width=\linewidth]{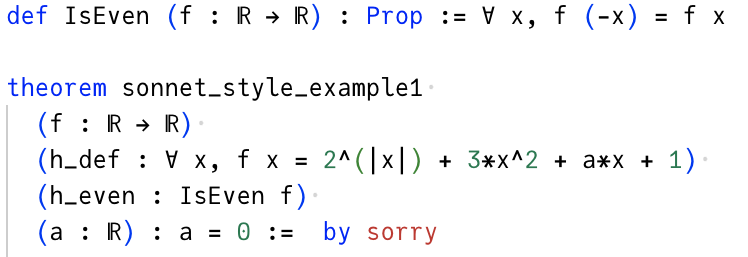}
\end{minipage} \newline
\texttt{Pass rate: 0/16} &
\begin{minipage}[t]{\linewidth}
    \includegraphics[width=0.8\linewidth]{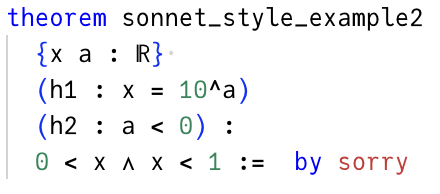}
\end{minipage} \newline
\texttt{Pass rate: 5/16} \\ \hline

\end{tabular}
}
\caption{Comparison of formalizer outputs for two examples. In Example 1, Formalizer A defines the "even function" directly by stating \( f(-x) = f(x) \). In contrast, Formalizer B first introduces a function called "IsEven" and then defines the even function using "IsEven". Notably, our prover successfully solves the statements provided by Formalizer A but fails with those from Formalizer B.
Example 2 is similar; however, our prover fails to solve the statement provided by Formalizer A but succeeds with the one from Formalizer B.}
\label{tab:formalizer_comparison_two_columns}
\end{table}

\subsection{Quality Assessment Details}
\label{app:qual_assess}
\begin{figure}
    \centering
    \includegraphics[width=0.9\linewidth]{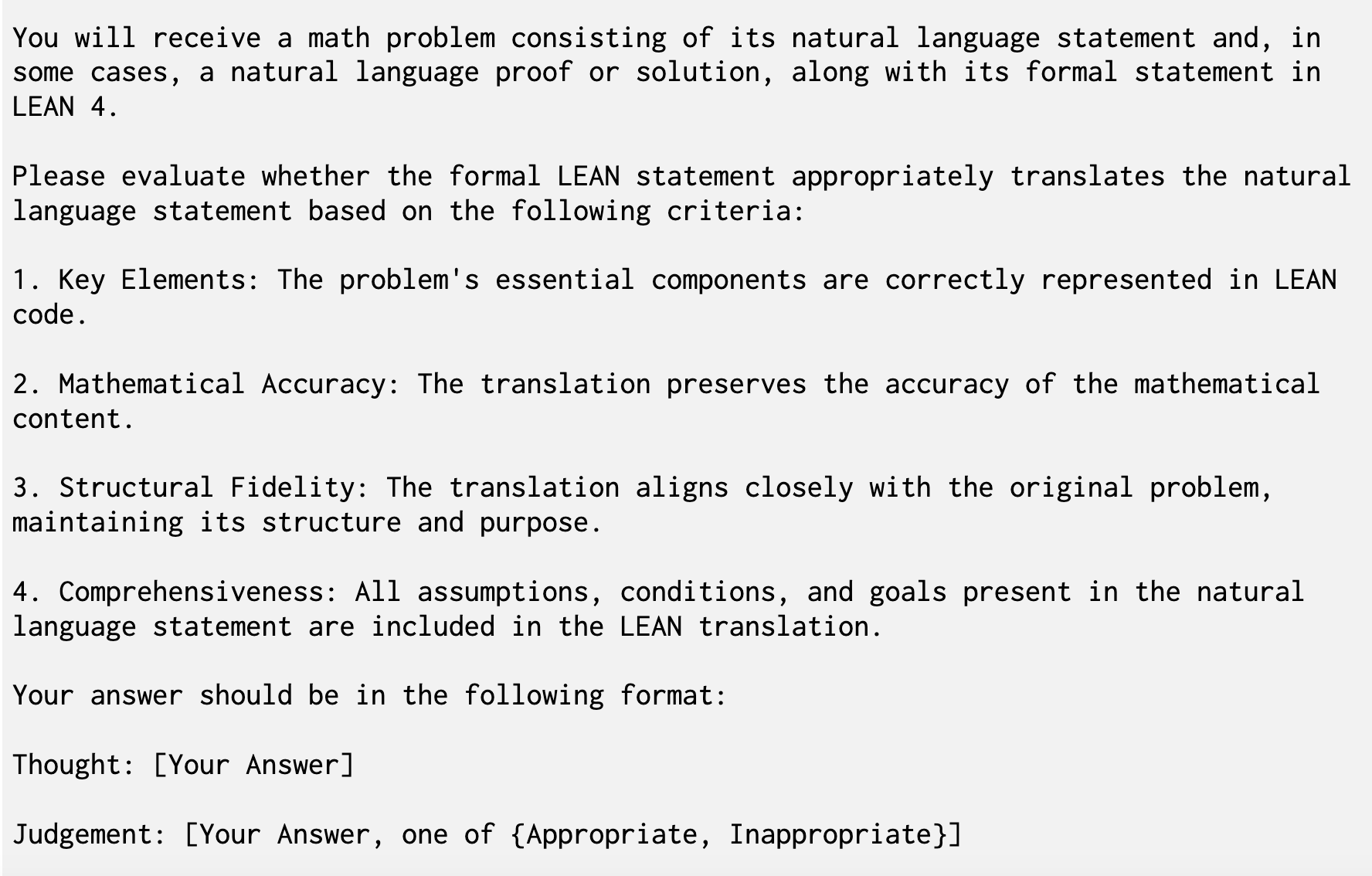}
    \caption{Prompts for Faithfulness and Completeness (FC) Test.}
    \label{fig:prompt:assessment}
\end{figure}

For the FC test, we use \qweninstruct{}\footnote{\url{https://huggingface.co/Qwen/Qwen2.5-72B-Instruct}} with prompt shown in
Figure~\ref{fig:prompt:assessment}. For each formalized statement, we generate four independent judgments, and the FC score is calculated as \#\{``Appropriate'' in four Judgments\}/4. For example, if the four judgments produced by \qweninstruct{} include three ``Appropriate'' and one ``Inappropriate'', the overall FC score is calculated as 0.75. We filter out formalized statements with an FC score less than 0.5.

\begin{table}[t]
    \centering
    \begin{tabular}{c|ccc}
        \toprule
        Model & Pass & Formalizer A & Formalizer B\\
        \midrule
        CC Test &Pass@1&  76.74\% & 88.48\% \\
        CC Test &Pass@8& 95.93\% & 98.59\%\\
        \midrule
        FC Test &Pass@1& 48.06\% & 80.42\%\\
        FC Test &Pass@8& 88.01\% & 97.22\%\\
        \midrule
        CC + FC Test &Pass@1& 45.72\% & 76.41\%\\
        CC + FC Test &Pass@8& 82.33\% & 95.78\%\\
         \bottomrule
    \end{tabular}
    \caption{Quality assessment of the formalized statement}
    \label{tab:formalization_statement_assessment}
\end{table}

For each informal statement in Numina, we generate eight formalized statements from each formalizer, resulting in 16 formalized statements per problem. Each statement undergoes the CC and FC Test, and we retain only those valid statements. We then randomly select one valid statement from each formalizer. For example, if five out of eight statements from Formalizer A and three from Formalizer B are valid, we randomly choose one from each. If a formalizer produces no valid statements, we exclude all its statements for that problem. The statistics for each test conducted on both formalizers are summarized in Table~\ref{tab:formalization_statement_assessment}.

\section{Expert Iteration Details}
\label{app:training_details}


The main training pipeline is illustrated in Section~\ref{sect:iterative}. When we implement the expert iteration algorithm, we gradually add the data. From iter-0 to iter-3, we gradually add the statements formalized by Claude-sonnet-3.5. At iter-3, we train the Formalizer B and add the formalized statements generated by Formalizer B for iter-4 to iter-6. At iter-7, we begin to add the statements generated by Formalizer A. We also add \mathlib{} data into the training dataset for better ProofNet performance when starting from iter-6.

\begin{table}[h]
    \centering
    \resizebox{\linewidth}{!}{
    \begin{tabular}{c|cc|ccc}
    \toprule
         \multirow{2}{*}{Iteration} & \multicolumn{2}{c|}{Statements} & \multicolumn{3}{c}{Training Data}\\
       & \lwb{} & Formalized & \lwb{} Solved & Formalized Solved & \mathlib{} \\
      \midrule
       Iter-0  & 140K & 0 & 20.6K & 0 & 0\\
       Iter-1  & 140K & 140K & 20.6K & 72.4K & 0\\
       Iter-2  & 140K & 270K & 23.0K & 128.7K & 0\\
       Iter-3  & 140K & 270K & 24.4K & 161.2K & 0 \\
       Iter-4  & 140K & 882K & 25.4K & 425.8K & 0\\
       Iter-5  & 140K & 882K & 27.0K & 436.5K & 0\\
       Iter-6  & 140K & 882K & 27.8K & 443.2K & 104K\\
       Iter-7  & 140K & 1.64M & 28.8K & 887.7K & 104K\\
       Iter-8  & 140K & 1.64M & 29.7K & 915.7K & 104K\\
       Iter-9  & 140K & 1.64M & 30.3K & 928.2K & 104K\\
    \bottomrule
    \end{tabular}
    }
    \caption{Expert iteration details.}
    \label{tab:data_for_training}
\end{table}

\section{More examples on style difference}
\label{app:examples}
\subsection{\mathlib{} and \miniff{}}
\label{app:mathlib_examples}
\begin{figure}[h]
    \centering
    \includegraphics[width=0.8\linewidth]{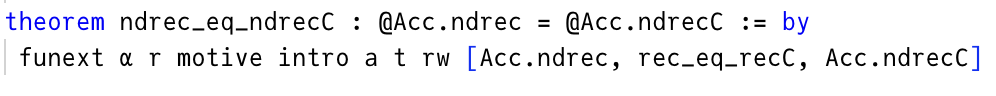}
    \caption{A \mathlib{} example which relies on pre-defined objects \texttt{@Acc.ndrec} and \texttt{@Acc.ndrecC}}
    \label{fig:mathlib4_example}
\end{figure}

\begin{figure}[h]
    \centering
    \includegraphics[width=0.8\linewidth]{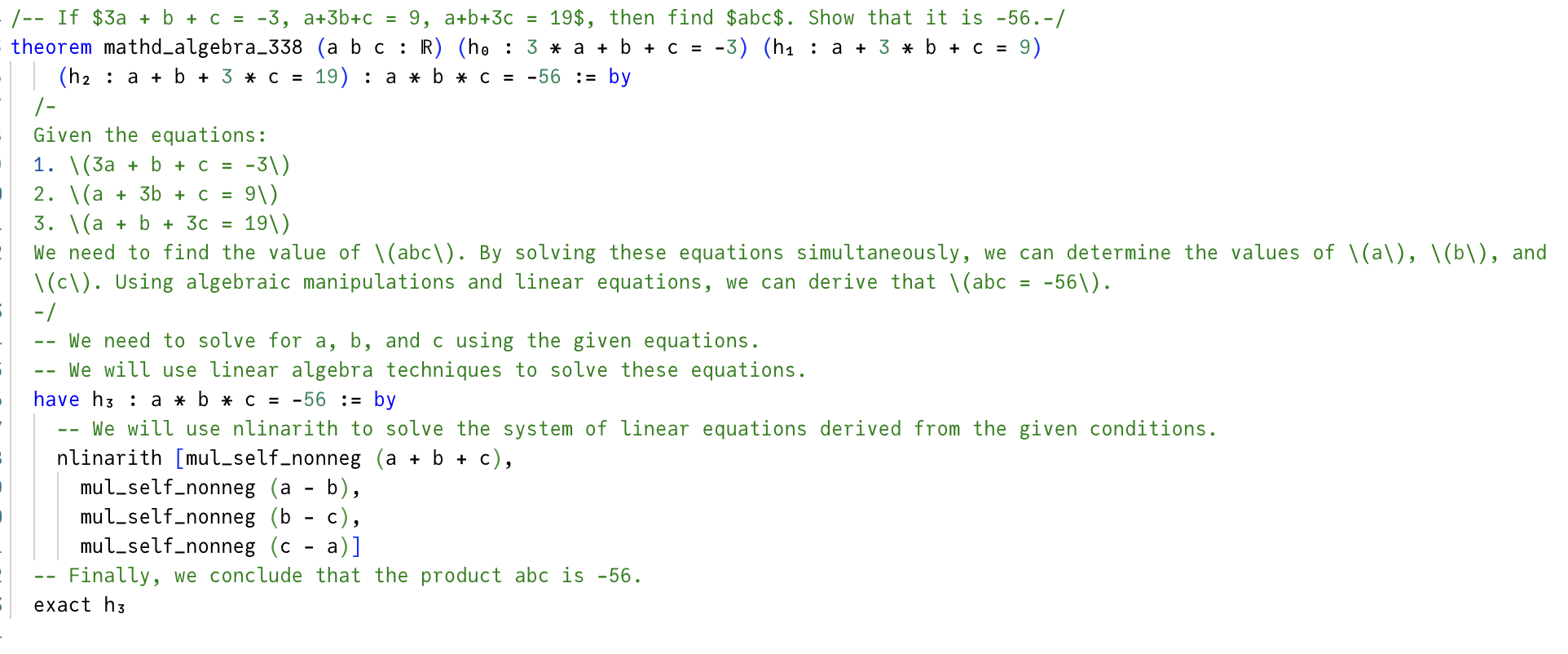}
    \caption{A \miniff{} example which does not rely on pre-defined objects}
    \label{fig:minif2f_example2}
\end{figure}

We observe a notable difference in the distribution of \mathlib{} compared to that of general problem-solving benchmarks, such as the widely used miniF2F~\citep{zheng2021minif2f}. For instance, \miniff{} largely consist of competition and Olympic-style problems, which require complex reasoning, while only depending on a relatively small set of elementary facts about integers, real numbers, counting, and geometry. On the contrary the statements in \mathlib{} focus on the simple manipulation of advanced mathematical concepts. Figure \ref{fig:mathlib4_example} and \ref{fig:minif2f_example2} show the statement and proof in \mathlib{} and \miniff{} respectively. It can be easily seen that both the statement and proof rely on pre-defined objects. Unlike \miniff{} statements, the example in Figure \ref{fig:mathlib4_example} can not even pass the lean compilation, given that pre-defined objects are missing.

\subsection{\proofnet{} and \miniff{}}
\label{app:proofnet_examples}
The problems in \proofnet{} are primarily drawn from undergraduate pure mathematics textbooks, covering topics such as real and complex analysis, linear algebra, abstract algebra, and topology. These topics largely rely on the abstract and general formulations of mathematical definitions in \mathlib{} \citep{mathlib4}.  We show two examples in Table \ref{tab:proofnet_miniF2F_comparison} to illustrate the style difference between \proofnet{} and \miniff{}.

\begin{table}[t]
\centering
\renewcommand{\arraystretch}{1.5} 
\begin{tabular}{|p{0.15\textwidth}|p{0.35\textwidth}|p{0.35\textwidth}|}
\hline
 & \textbf{Example from \proofnet{}} & \textbf{Example from \miniff{}} \\ \hline

\textbf{Informal Statement} &
Prove that no order can be defined in the complex field that turns it into an ordered field. &
Show that for any natural number $n$, $7$ does not divide $2^n + 1$. \\ \hline

\textbf{Formal Statement} &
\begin{minipage}[t]{\linewidth}
    \includegraphics[width=\linewidth]{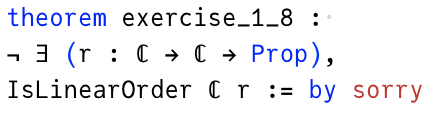}
\end{minipage} &
\begin{minipage}[t]{\linewidth}
    \includegraphics[width=\linewidth]{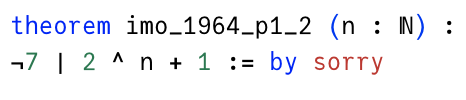}
\end{minipage} \\ \hline

\textbf{Comments} &
This problem involves the notion of order, which is undergraduate level. Its formal statement uses the definition \texttt{IsLinearOrder} in \mathlib{}. &
This problem comes from IMO but only involves division. \\ \hline

\end{tabular}
\caption{Comparison of Examples from \proofnet{} and \miniff. \proofnet{} largely relies on the abstract and general formulations of mathematical results in \mathlib{}. In contrast, \miniff{} largely consists of high-school competition and Olympic style problems, which require complex reasoning. }
\label{tab:proofnet_miniF2F_comparison}
\end{table}

\section{Alternative approach for synthesizing data}
\label{app:sketch_proof}
We also considered other pipeline beyond autoformalizing statement and expert iteration for collecting proof data. Inspired by \cite{jiang2022draft}, we implemented the following pipeline: 
\paragraph{Step 1.} We prompt OpenAI’s o1-preview model to generate a proof for a formal statement. We ask the model to generate the proof step-by-step, use \texttt{"have"} tactic to structure the proof. For each proof step, the subgoal of this step is indicated by \texttt{"have"}, following by proofs for this subgoal.
\paragraph{Step 2.} We remove the proofs for the subgoal provided by o1-preview in each \texttt{"have"} block (these proofs often involves detailed lean syntax, and is usually incorrect). That is, we only keep the "sketch" of the proof. We then put this proof sketch into Lean compiler, to automatically extract each subgoal and corresponding conditions, to form several subproblems.
\paragraph{Step 3. } We apply \dsprover-RL to try to proof the subproblems. We try each subproblem for 32 times. If all subproblems are successfully proved, assembling these subproofs into the sketch gives us a valid proof for the original problem. 

\begin{figure}[h]
    \centering
    \includegraphics[width=0.8\linewidth]{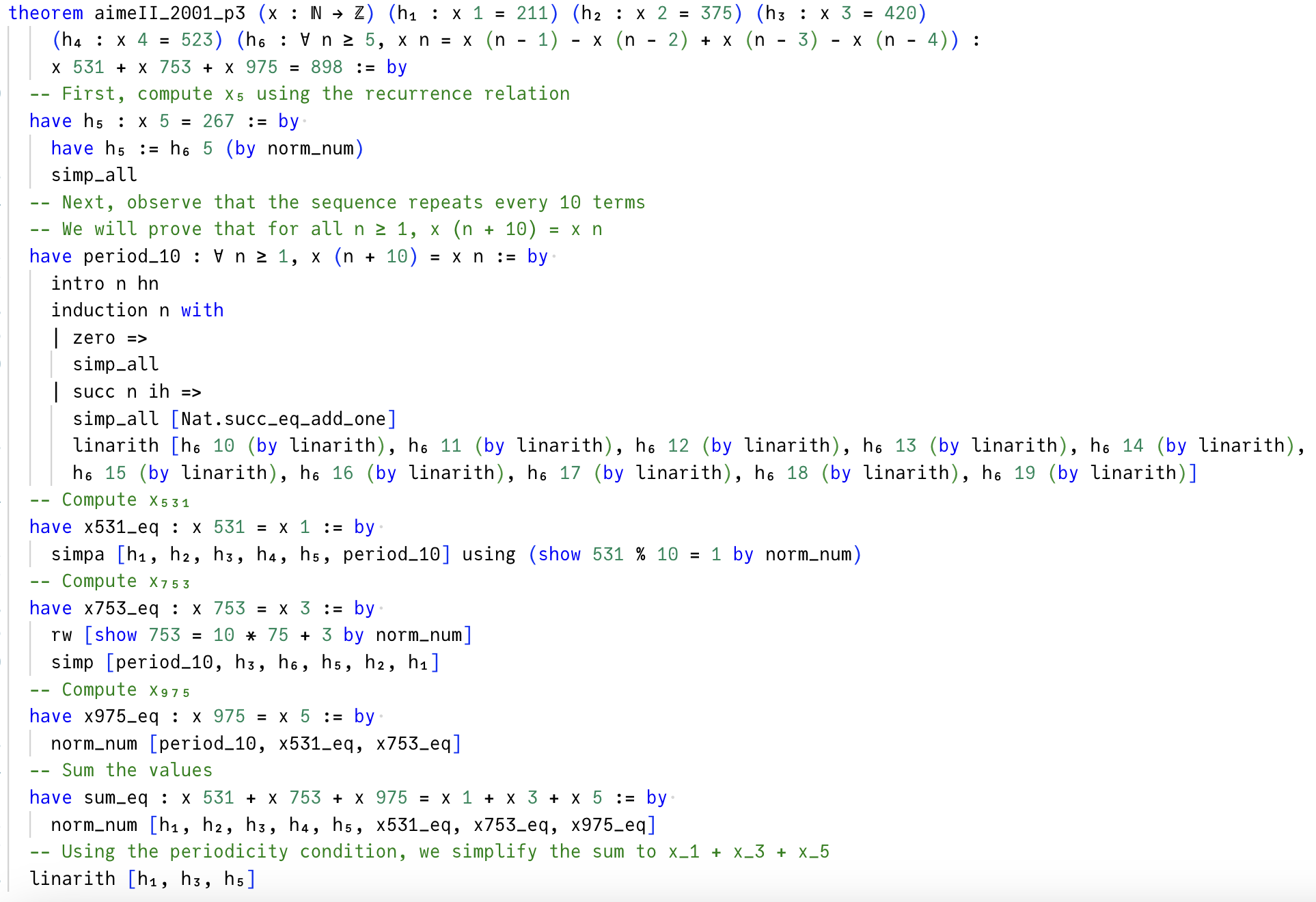}
    \caption{A non trivial problem solved by the divide-and-conquer pipeline}
\label{fig:sketch_proof_example}
\end{figure}

Figure \ref{fig:sketch_proof_example} shows the only problem solved by this pipeline that \dsprover-RL does not solve, which is a non trivial problem that requires relatively complex reasoning. Though this pipeline has shown potential, the efficiency is quite low. Only one additional problem is proved using this pipeline, among 244 problems in \miniff{} validation set. This might due to the fact that this pipeline is overly complicated, since failure of each subproblem might lead to the failure for the entire problem.

\section{RL training details}
\label{app:RL}
\subsection{DPO training}
For DPO training, we construct pairwise data on problems with pass ratio in $(0,1/4]$ (from previous training dataset). To be specific, for each problem, we do Pass@16, and the pass ratio $(0,1/4]$ means we select samples where \prover-SFT generates 1-4 correct proofs within 16 trails. We construct DPO pairs by randomly select a correct proof and wrong proof from the 16 trials. We sample 508K proved problems from the original dataset, and among which 30K problems with the aforementioned pass ratio is selected. We use a learning rate of $5 \times 10^{-6}$ and train for two epoches.

\begin{figure}[h]
    \centering
    \includegraphics[width=0.8\linewidth]{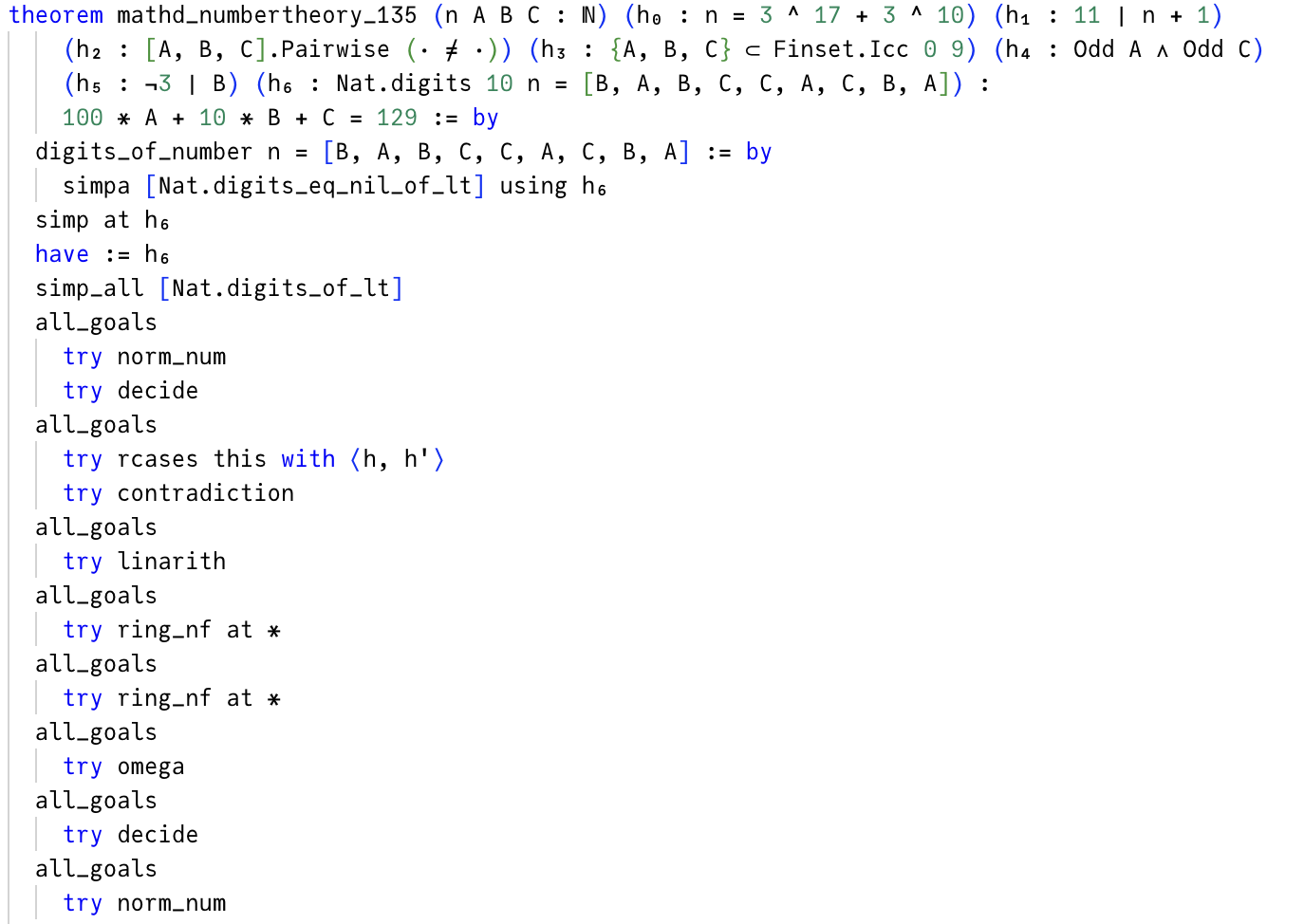}
    \caption{Example of output of DPO model. The model is repeatedly using \texttt{all\_goals} and \texttt{try}. }
    \label{fig:dpo_example}
\end{figure}

Our experiments reveal that through DPO training, the model is easy to learn "shortcuts". Figure \ref{fig:dpo_example} shows one typical output of the DPO model. It repeatedly use tactics \texttt{all\_goals} and \texttt{try}, which might be shortcuts learned in DPO training. To mitigate the model's tendency to produce verbose, lengthy proofs by repeatedly utilizing these shortcuts, we implement length regularization in our DPO framework. Specifically, when multiple correct answers are available for a given statement, we select the one with the shortest length. All other settings remain unchanged from the original DPO implementation.

\subsection{GRPO training}

We collect 80K problem statements whose pass ratio is within $(0,1/2]$. We will also explore different design choices for the included problems in the subsequent discussion. Using these problem statements, we employed the \prover-SFT as our base model and conducted reinforcement learning (RL) training within the OpenRLHF framework, utilizing the GRPO algorithm.  During the RL training, we generated 16 proofs for each problem and verified their correctness through compilation. Correct proofs received a reward of +8, while incorrect proofs received a penalty of -8. We search for the learning rate among $1 \times 10^{-5}$, $5\times10^{-6}$, $2\times10^{-6}$, and $1\times10^{-6}$ and choose the learning rate $5\times10^{-6}$. We explored initial KL penalty values of 0.03, 0.003, 0.00003, and 0. Our findings indicate that the KL penalty does not significantly impact training. Consequently, we selected 0.003 as the penalty weight.We used a batch size of 256 and also tested a batch size of 128, which achieved very similar performance. After training the RL model for one epoch, we found that increasing the number of epochs does not enhance the final testing accuracy.

\begin{figure}[h]
    \centering
    \includegraphics[width=0.8\linewidth]{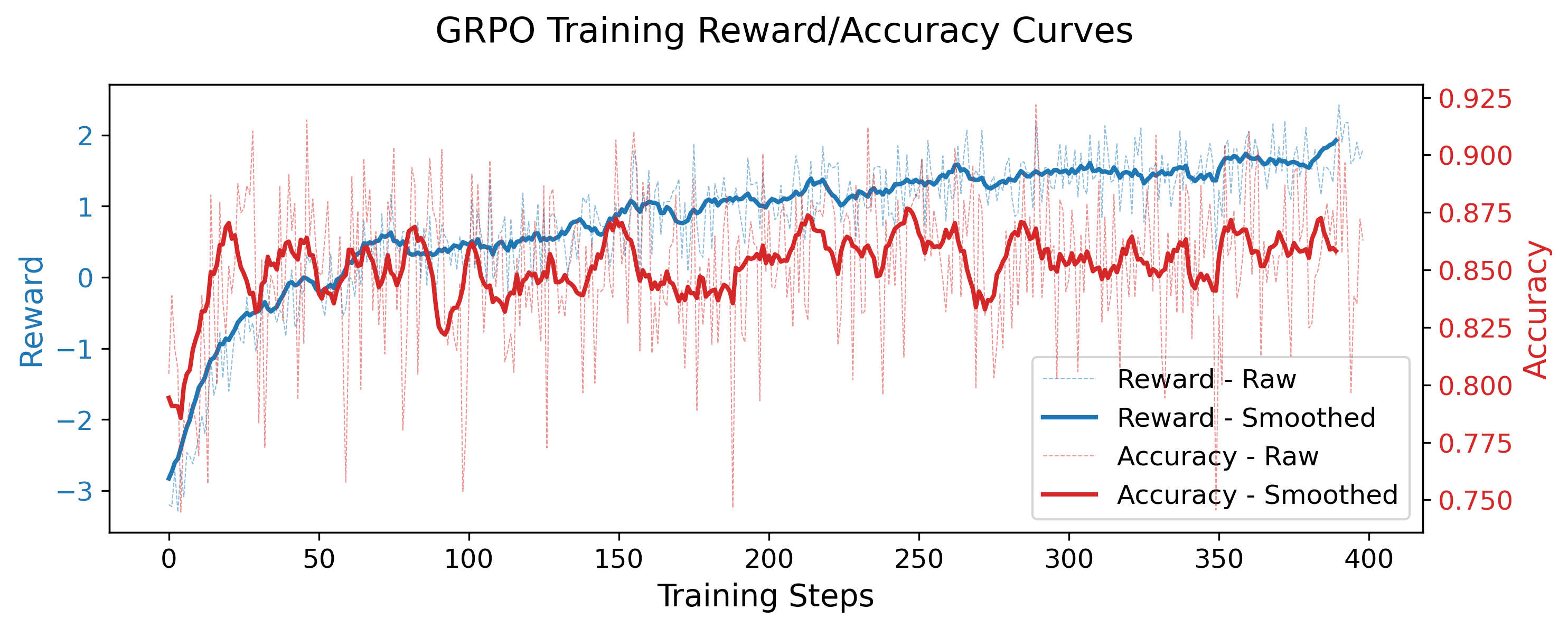}
    \caption{This figure illustrates the average reward/accuracy of each batch during GRPO training. A correct proof corresponds to a reward +8, while failed one has a reward -8.}
    \label{fig:RL_training_curves}
\end{figure}

\paragraph{Mismatch between reinforcement learning (RL) reward and test accuracy.}
Figure~\ref{fig:RL_training_curves} plots the average training reward and Pass@16 accuracy across training batches. Notably, we observe a mismatch between the reward and accuracy trends: while the average reward continues to increase throughout training, the Pass@16 accuracy plateaus after approximately 20 training steps. This discrepancy may stem from the misalignment between the optimization objective and the evaluation metric. GRPO encourages generating successful proofs more frequently, rewarding higher success rates across samples. In contrast, the Pass@N metric only considers whether a problem is solved at least once, irrespective of how many successful attempts occur. As a result, improvements in reward do not necessarily translate into better Pass@N performance.

\paragraph{Exploration of included prompts for training RL.} 
We previously mentioned that we use statements with a pass ratio within \((0, 1/2]\) for training the RL model. This selection is based on the fact that these samples are challenging yet manageable for the current checkpoint. We also conducted experiments with pass ratios of \((0, 1/4]\), \((0, 3/4]\), and \((0, 1]\). Our findings indicate that balancing the difficulty of the chosen prompts is crucial, and we compared their performance in terms of final testing results in Table~\ref{tab:included_prompt_sets}.

\begin{table}[]
    \centering
    \begin{tabular}{c|c|c}
    \toprule
        Prompt Pass Ratio & Prompt Number & mini-F2F Accuracy(\%) \\
        \midrule
        (0, 1/4] & 30K&  58.2\\
        (0, 1/2] & 62K& \textbf{60.4}\\ 
        (0, 3/4] & 115K& 59.8\\
        (0, 1]  & (sub-sample) 200K&59.2\\
        \bottomrule
    \end{tabular}
    \caption{Results of included different prompts for training RL.}
    \label{tab:included_prompt_sets}
\end{table}

\paragraph{Exploration on the reward design for timeout samples.} Typically, when using the Lean compiler to verify a Lean proof, we encounter three possible outcomes: successful compilation, failure with returned errors, or a timeout within the predefined time limit. We experiment with various rewards for the timeout samples, while maintaining a fixed reward of +8 for correct proofs that compile successfully and -8 for incorrect proofs that fail to compile. The results in Table~\ref{tab:time_out} demonstrates that setting the reward for timeouts to be the same as that for failures results in improved performance across these experiments.

\begin{table}[]
    \centering
    \begin{tabular}{c|cc}
    \toprule
        Timeout Reward & Testing Timeout Ratio & Testing Accuracy \\
        \midrule
        0 & 4.5\%& 58.7\%\\
        -8 & 1.7\% & 60.2\%\\
        -16 & 0.8\% & 59.2\%\\
         \bottomrule
    \end{tabular}
    \caption{Investigation on the reward for timeout samples}
    \label{tab:time_out}
\end{table}

\section{Discussion}
\label{app:discussion}

We delve into
the characteristics of proofs generated by \prover-SFT and discuss potential directions for improvement, particularly regarding the proof style adopted by the model, the role of search as well as online interaction in proof generation, and the integration of external symbolic computation tools such as SymPy.

\begin{figure}[h]
    \centering
    \includegraphics[width=0.8\linewidth]{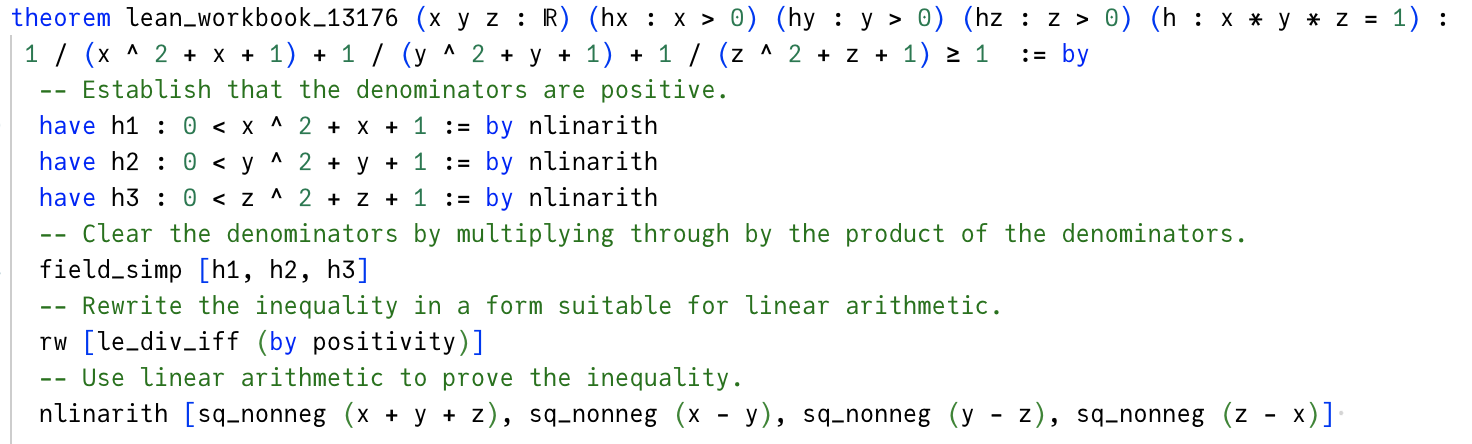}
    \caption{Example of proof style, where intermediate steps are absorbed in high-level tactics}
    \label{fig:proof_style}
\end{figure}

\paragraph{The Proof Style.}
We observe that the proofs provided by \prover-SFT often rely on high-level tactics such as \texttt{nlinarith} and \texttt{simp\_all} among others. These high-level tactics handle multiple reasoning steps internally, delegating the resolution of intermediate steps to their built-in automation. For example, the \texttt{nlinarith} tactic can automatically solve certain linear and non-linear equalities and inequalities.
Figure \ref{fig:proof_style} shows a typical proof generated by our prover. The first several steps involve only trivial transformations of the original problem, whereas the final line uses \texttt{nlinarith} to immediately achieve the goal. Whether this style of proof is sufficient for complex reasoning remains an important area for exploration.

\paragraph{Search and online interaction.}
Currently, \prover-SFT generates the entire proof for the problem at once, without receiving further feedback. While our current approach is appealing in terms of computation, incorporating search and interaction in future work could enhance performance. For example, once a tactic is generated by our prover, it can interact with the Lean compiler to receive feedback on how the goal changes after the tactic is applied. This information can then be utilized in generating the next tactic, potentially improving the overall proof strategy~\citep{wu2024internlm2}. 

\paragraph{SymPy.} 
Future work may aim to leverage other software packages to enhance Lean's capabilities. For instance, Lean's \texttt{ring} tactic can handle algebraic simplifications by applying axioms such as distributivity, associativity, and commutativity. However, a combination of tactics is required for non-algebraic transformations of transcendental functions, such as logarithmic and trigonometric functions, and other advanced simplifications beyond commutative rings. We explored using a Python-based computer algebra system, SymPy~\citep{meurer2017sympy}, to simplify complex expressions in theorem statements and feed the simplified form into the prover.
Specifically, we parse equations of the form \( A = B \) within the goals of Lean theorem statements, construct the SymPy expression \( A - B \), and then apply the \texttt{simplify} method in Lean. This procedure directly solves 9.4\% of \miniff{} by simplifying the statements to $0=0$. In addition, it solves 0.8\% of the problems in \miniff{} that were unsolved by \prover-SFT with Pass@32, but did not improve \prover-SFT with Pass@3200. Thus, SymPy simplification is not part of any of our reported results. However, we think such procedures need further exploration. 

\end{document}